\documentclass[runningheads]{llncs}
\usepackage[T1]{fontenc}
\usepackage{graphicx}
\usepackage{booktabs}
\usepackage[misc]{ifsym}

\usepackage{bm}
\usepackage{multirow} 
\usepackage{pifont}
\usepackage{makecell}
\usepackage{colortbl}
\usepackage{xcolor}
\usepackage{amsmath}
\usepackage{amssymb}
\usepackage{subcaption}
\usepackage{microtype}
\usepackage[misc]{ifsym}

\newcommand{\psm}[1]{\scalebox{0.4}{$\pm$#1}}

\begin{document}

\tocauthor{Rongzhen Zhao, Vivienne Wang, Juho Kannala, Joni Pajarinen}
\toctitle{Grouped Discrete Representation for Object-Centric Learning}

\title{Grouped Discrete Representation for Object-Centric Learning}


\author{Rongzhen Zhao$^{1}$ \Letter\and Vivienne Wang$^{1}$\and Juho Kannala$^{2,3}$\and Joni Pajarinen$^{1}$\\[1ex]
}


\institute{Department of Electrical Engineering and Automation, Aalto University, Finland \email{\{rongzhen.zhao, vivienne.wang, joni.pajarinen\}@aalto.fi}
\and
Department of Computer Science, Aalto University, Finland \email{juho.kannala@aalto.fi}
\and
Center for Machine Vision and Signal Analysis, University of Oulu, Finland
}

\maketitle              

\begin{abstract}
Object-Centric Learning (OCL) aims to discover objects in images or videos by reconstructing the input.
Representative methods achieve this by reconstructing the input as its Variational Autoencoder (VAE) discrete representations, which suppress (super-)pixel noise and enhance object separability.
However, these methods treat features as indivisible units, overlooking their compositional attributes, and discretize features via scalar code indexes, losing attribute-level similarities and differences.
We propose Grouped Discrete Representation (GDR) for OCL. For better generalization, features are decomposed into combinatorial attributes by organized channel grouping. For better convergence, features are quantized into discrete representations via tuple code indexes.
Experiments demonstrate that GDR consistently improves both mainstream and state-of-the-art OCL methods across various datasets.
Visualizations further highlight GDR's superior object separability and interpretability.
The source code is available on https://github.com/Genera1Z/GroupedDiscreteRepresentation.

\end{abstract}

\section{Introduction}
\label{sect:introduction}

Under self or weak supervision, Object-Centric Learning (OCL) \cite{greff2019iodine,burgess2019monet} represents dense image or video pixels as sparse object feature vectors, known as \textit{slots}. These slots can be used for \textit{set prediction} while their corresponding attention maps for \textit{object discovery} \cite{locatello2020slotattent}. OCL is bio-plausible, as humans perceive visual scenes as objects for visual cognition, like understanding, reasoning, planning, and decision-making \cite{bar2004visual,cavanagh2011visual,palmeri2004visual}. OCL is versatile, as object-level representations of images or videos fit to tasks involving different modalities \cite{yi2019clevrer,wu2022slotformer}.

The training signal comes from reconstructing the input.
Directly reconstructing input pixels \cite{locatello2020slotattent,greff2019iodine} struggles with complex-textured objects.
Mixture-based OCL methods \cite{kipf2021savi,elsayed2022savipp} reconstruct more object-separable modalities, like optical flow and depth maps.
Foundation-based methods \cite{seitzer2023dinosaur,zadaianchuk2024videosaur} use the input's foundation model features as the target.
Transformer-based \cite{singh2021slate,singh2022steve} and Diffusion-based methods \cite{wu2023slotdiffuz,jiang2023lsd} reconstruct the input's Variational Autoencoder (VAE) intermediate representation. With a limited number of shared template features, i.e., codes in a codebook, continuous-valued superpixels in VAE representations are discretized \cite{im2017dvae,van2017vqvae}. This suppresses (super-)pixel noise and enhances object separability.
Empirically, improved object separability in the reconstruction target offers OCL more effective training guidance.

However, these methods treat features as atomic units and entangle their composing attributes together, thus limiting model generalization.
Moreover, the corresponding scalar code indexes fail to capture superpixels' attribute-level similarities and differences, thus hindering model convergence.

\begin{figure}[t]
\includegraphics[width=\textwidth]{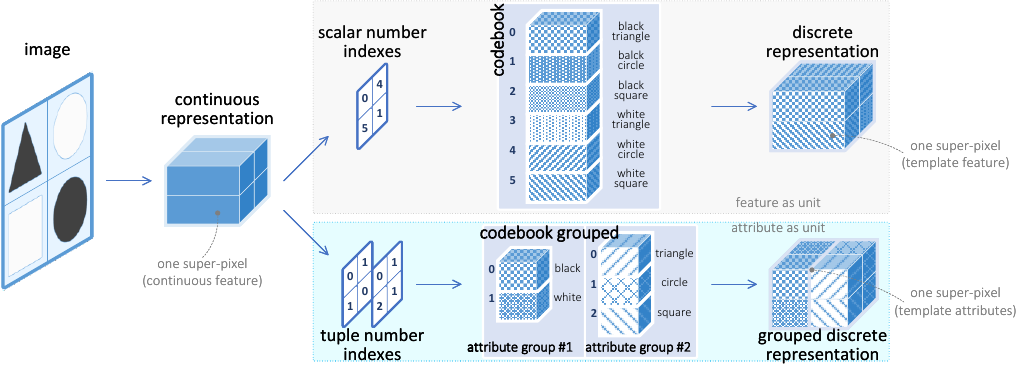}
\centering
\caption{
Non-grouped vs grouped discrete representation. (\textit{upper}) Existing methods treat features as units, selecting template features from a codebook by scalar indexes to discretize superpixels. (\textit{lower}) We treat attributes as units, selecting template attributes from a grouped codebook by tuple indexes.
}
\label{fig:teaser}
\end{figure}

As illustrated in Fig.~\ref{fig:teaser}, consider a dataset characterized by two attribute groups: color (black, white) and shape (triangle, square, circle). An image in it contains four objects, each downsampled to a superpixel in the feature map.
To select template features from a feature-level codebook, six scalar code indexes are needed, where digits 0-5 refer to black-triangle, black-circle, black-square, etc. Each code is reused with probability $\frac{1}{6}$. The feature map can thus be discretized as $[\begin{smallmatrix}0 & 4\\5 & 1\end{smallmatrix}]$.
But if decomposed, superpixels can be discretized as combinations of template attributes from two attribute groups, i.e., $[\begin{smallmatrix}0,0&1,1\\1,2&0,1\end{smallmatrix}]$. 
The first and second numbers in these index tuples indicate whether superpixels' attributes are the same or different, facilitating model convergence. 
These codes are reused with higher probabilities $\frac{1}{2}$ and $\frac{1}{3}$ respectively, benefiting model generalization.

Our main contributions are as follows:
(\textit{i}) We propose \textit{Grouped Discrete Representation} (GDR) for VAE discrete representation to guide OCL training better;
(\textit{ii}) GDR is compatible with mainstream OCL methods and boosts both their convergence and generalization;
(\textit{iii}) GDR captures attribute-level similarities and differences, also enhances object separability in VAE representations.

\section{Related Work}
\label{sect:related_work}

\textbf{Object-Centric Learning} (OCL).
Mainstream OCL obtains supervision from reconstruction using slots aggregated by SlotAttention \cite{locatello2020slotattent,bahdanau2014qkvattent} from the input's dense superpixels.
SLATE \cite{singh2021slate} and STEVE \cite{singh2022steve}, which are Transformer-based, generate input tokens from slots via a Transformer decoder \cite{vaswani2017transformer}, guided by dVAE \cite{im2017dvae} discrete representations.
SlotDiffusion \cite{wu2023slotdiffuz} and LSD \cite{jiang2023lsd}, which are Diffusion-based, recover input noise from slots via a Diffusion model \cite{rombach2022ldm}, guided by VQ-VAE \cite{van2017vqvae} discrete representations.
DINOSAUR \cite{seitzer2023dinosaur} and VideoSAUR \cite{zadaianchuk2024videosaur}, which are foundation model-based, reconstruct input features from slots via a spatial broadcast decoder \cite{watters2019sbd}, guided by well-pretrained features of the foundation model DINO \cite{caron2021dino1,oquab2023dino2}.
We focus on the VAE part of OCL.

\textbf{Variational Autoencoder} (VAE).
Discrete representations of VAE have been shown to guide OCL better than direct input pixels as reconstruction targets.
Transformer-based OCL methods \cite{singh2021slate,singh2022steve} utilize dVAE \cite{im2017dvae} to discretize encoder representations by selecting template features from a codebook via Gumbel sampling \cite{jang2016gumbelsoftmax}.
Diffusion-based OCL methods \cite{jiang2023lsd,wu2023slotdiffuz} employ VQ-VAE \cite{van2017vqvae} to achieve discretization by replacing features with their closest codebook codes.
Similar to our idea, both \cite{bouchacourt2018mlvae} and \cite{liu2023cdisentangle} seek to decompose features into attributes, but their monolithic VAE representation is incompatible with OCL.
Other VAE variants also offer techniques, like grouping \cite{yang2023hificodec}, residual \cite{barnes1996residualvq} and clustering \cite{lim2020clustervae}, to enhance VAE representations.
We borrow some for the OCL setting.

\textbf{Channel Grouping}.
Splitting features along the channel dimension and transforming them separately is often used to diversify representations \cite{krizhevsky2012alexnet,chen2019octconv,huang2018condensenet,zhao2021lhc,zhao2022coc}. 
These solutions mainly perform grouping directly on feature maps \cite{krizhevsky2012alexnet} or on learnable parameters \cite{zhao2022coc}.
To the best of our knowledge, only one work has explored this idea in the OCL setting.
SysBinder \cite{singh2022sysbind} groups slots along the channel dimension in the slot attention \cite{locatello2020slotattent} to aggregate different attributes of objects, yielding better interpretability in object representation but limited performance gains.
We group VAE intermediate representations along channels, yielding grouped discrete representations to guide OCL training better.

\begin{figure}[t]
\centering
\includegraphics[width=\textwidth]{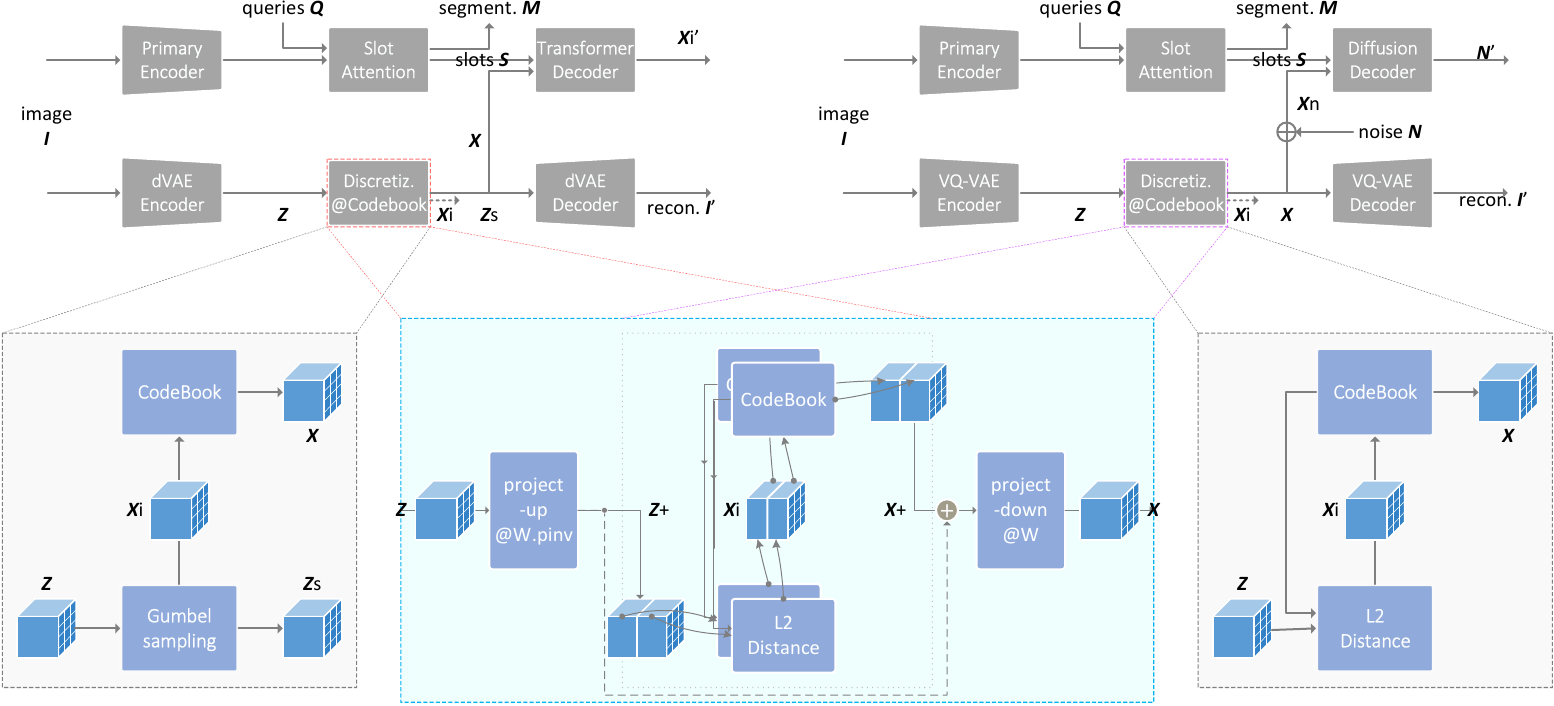}
\caption{
Our GDR is applicable to mainstream OCL.
First row: architectures of Transformer-based (\textit{left}) and Diffusion-based (\textit{right}) methods.
Second row: non-grouped representation discretization in dVAE (\textit{left}), non-grouped discretization in VQ-VAE (\textit{right}), and grouped discretization (\textit{center}) of our method.
}
\label{fig:preliminary}
\end{figure}

\section{Proposed Method}
\label{sect:proposed_method}

We propose Grouped Discrete Representation (GDR), applicable to mainstream OCL methods, either Transformer-based \cite{singh2021slate,singh2022steve,kakogeorgiou2024spot,zadaianchuk2024videosaur} or Diffusion-based \cite{wu2023slotdiffuz,jiang2023lsd}.
Simply modifying their VAE, our GDR improves them by providing reconstruction targets, or \textit{guidance}, with better object separability.

\textbf{Notations}: As shown in Fig. \ref{fig:preliminary}, image or video frame $\bm{I}$, continuous representation $\bm{Z}$, discrete representation $\bm{X}$, and noise $\bm{N}$ are tensors in shape (height, width, channel); queries $\bm{Q}$ and slots $\bm{S}$ are tensors in shape (number-of-slots, channel); segmentation $\bm{M}$ is a tensor in shape (height, width).

\subsection{Preliminary: Discrete Representation}
\label{sect:preliminary}

Both Transformer-based and Diffusion-based methods learn to aggregate pixels into \textit{slots} by reconstructing the input as its VAE discrete representation.

\textit{Transformer-based} architecture is depicted in Fig.~\ref{fig:preliminary} first row left. 
The input image or video frame $\bm{I}$ is encoded by a primary encoder and aggregated by SlotAttention \cite{locatello2020slotattent} into slots $\bm{S}$ under queries $\bm{Q}$, with object (and background) segmentation masks $\bm{M}$ as byproducts.
Meanwhile, pretrained VAE represents $\bm{I}$ as discrete representation $\bm{X}$ and the corresponding code indexes $\bm{X}_\mathrm{i}$.
Subsequently, using a Transformer decoder, $\bm{S}$ is tasked with reconstructing $\bm{X}_\mathrm{i}$ as classification, guided by causal-masked $\bm{X}$.
For videos, current slots $\bm{S}$ are transformed by a Transformer encoder block into queries for the next frame.

Specifically, discrete representations for Transformer-based OCL are obtained as shown in Fig.~\ref{fig:preliminary} second row left:
\begin{itemize}
\item Predefine a codebook $\bm{C}$ containing $n$ $c$-dimensional learnable codes as template features;
\item Transform input $\bm{I}$ with a dVAE encoder into continuous intermediate representation $\bm{Z}$;
\item Sample $\bm{Z}$ via Gumbel softmax, yielding one-hot indexes $\bm{X}_\mathrm{i}$ and soft sampling $\bm{Z}_\mathrm{s}$ for dVAE decoding;
\item Select template features from $\bm{C}$ by $\bm{X}_\mathrm{i}$ and compose the discrete representation $\bm{X}$ to guide OCL training.
\end{itemize}

\textit{Diffusion-based} architecture is drawn in Fig.~\ref{fig:preliminary} first row right.
The key difference is that, with a conditional Diffusion model decoder, $\bm{S}$ is tasked with reconstructing Gaussian noise $\bm{N}$ added to $\bm{X}$ as regression.

Specifically, discrete representations for Diffusion-based OCL are obtained as in Fig.~\ref{fig:preliminary} second row right:
\begin{itemize}
\item Predefine a codebook $\bm{C}$ containing $n$ learnable codes as template features;
\item Transform input $\bm{I}$ via VQ-VAE encoder into continuous representation $\bm{Z}$;
\item Find the closest codes' indexes $\bm{X}_\mathrm{i}$ in $\bm{C}$ for each superpixel in $\bm{Z}$;
\item Form discrete representation $\bm{X}$ by selecting $\bm{C}$ using $\bm{X}_{\mathrm{i}}$, for OCL training.
\end{itemize}

\textit{Remark}. These methods' features as discretization units overlooks the composing attributes, thus impeding generalization. Their scalars as code indexes loses sub-feature similarities and differences, thus hindering convergence.

\subsection{Naive Grouped Discrete Representation}
\label{sect:naive_gdr}

Our naive GDR decomposes features into attributes via direct channel grouping in VQ-VAE for both Transformer- and Diffusion-based methods.

\textit{Beforehand}, suppose a dataset is fully described by $n$ $c$-dimensional template features, which are further decomposed into $g$ attribute groups. Each group consists of $a$ $d$-dimensional template attributes, $n$ = $a ^ g$ and $c$ = $g \times d$. 
Thus, we predefine a set of attribute codebooks $\bm{C}=\{\bm{C}^{(1)}, \bm{C}^{(2)} ... \bm{C}^{(g)}\}$, whose parameters are in shape ($g$, $a$, $d$). The combinations of these codes are equivalent to the non-grouped feature-level codebook, whose parameters are in shape ($n$, $c$). 

\textit{Afterwards}, we transform the input $\bm{I}$ with a VAE encoder into continuous intermediate representation $\bm{Z}$.
In VQ-VAE, we sample distances between $\bm{Z}$ and $\bm{C}$ via Gumbel noise, yielding tuple code indexes $\bm{X}_{\mathrm{i}}$:
\begin{equation}
\label{eq:vqvae_dist}
\small
\bm{D} = \mathrm{l2}(\bm{Z}^{(1)}, \bm{C}^{(1)}) \circ \mathrm{l2}(\bm{Z}^{(2)}, \bm{C}^{(2)}) ... \circ \mathrm{l2}(\bm{Z}^{(g)}, \bm{C}^{(g)})
\end{equation}
\begin{equation}
\label{eq:vqvae_gumbel_sample}
\small
\bm{D}_{\mathrm{s}} = \mathrm{softmax}(\frac{\bm{D}^{(1)} + \bm{G}^{(1)}}{\tau}) \circ \mathrm{softmax}(\frac{\bm{D}^{(2)} + \bm{G}^{(2)}}{\tau}) ... \circ \mathrm{softmax}(\frac{\bm{D}^{(g)} + \bm{G}^{(g)}}{\tau})
\end{equation}
\begin{equation}
\label{eq:vqvae_code_index}
\small
\bm{X}_{\mathrm{i}} = \mathrm{argmin} (\bm{D}^{(1)}_{\mathrm{s}}) \circ \mathrm{argmin} (\bm{D}^{(2)}_{\mathrm{s}}) ... \circ \mathrm{argmin} (\bm{D}^{(g)}_{\mathrm{s}})
\end{equation}
where $\bm{Z}^{(1)}$...$\bm{Z}^{(g)}$ are channel groupings of $\bm{Z}$; $\bm{G}^{(1)}$...$\bm{G}^{(g)}$ are Gumbel noises; $\circ$ is channel concatenation; $\mathrm{l2}(\cdot , \cdot)$ denotes L2 distances between every vector pair in its two arguments; $\bm{D}_{\mathrm{s}}$ is soft Gumbel sampling of distances $\bm{D}$ between continuous representations and codes; $\mathrm{argmin}(\cdot)$ is along the code dimension. 
For our grouped VAE, multiple code indexes are selected from the attribute groups, forming ``tuple indexes''. In contrast, the non-grouped VAE selects only one index from a feature-level codebook, forming ``scalar indexes''.

\textit{Subsequently}, we select template attributes by $\bm{X}_{\mathrm{i}}$ from $\bm{C}$, forming grouped discrete representation $\bm{X}$, which is the target of Diffusion decoding:
\begin{equation}
\label{eq:codebook_select}
\small
\bm{X} = \mathrm{select} (\bm{C}^{(1)}, \bm{X} _ {\mathrm{i}} ^ {(1)}) \circ \mathrm{select} (\bm{C}^{(2)}, \bm{X} _ {\mathrm{i}} ^ {(2)}) ... \circ \mathrm{select} (\bm{C}^{(g)}, \bm{X} _ {\mathrm{i}} ^ {(g)})
\end{equation}
where $\mathrm{index}(\cdot, \cdot)$ selects codes from a codebook given indexes.

\textit{Finally}, we transform $\bm{X}_{\mathrm{i}}$ from tuple into scalar, which is the target of Transformer decoding:
\begin{equation}
\label{eq:index_tuple_to_scalar}
\small
\bm{X}_{\mathrm{i}} := a^0 \times \bm{X}^{(1)}_{\mathrm{i}} + a^1 \times \bm{X}^{(2)}_{\mathrm{i}} + ... + a^{g-1} \times \bm{X}^{(g)}_{\mathrm{i}}
\end{equation}
where $\bm{X} _ {\mathrm{i}} ^ {(1)}$...$\bm{X} _ {\mathrm{i}} ^ {(g)}$ are the channel groupings of $\bm{X} _ {\mathrm{i}}$ from Eq. \ref{eq:vqvae_code_index}.

\textit{Besides}, we introduce a mild loss to encourage code utilization after grouping
\begin{equation}
\label{eq:utiliz_loss}
\small
l_{\mathrm{u}} = - \mathrm{entropy}(\mathbb{E}[D^{(1)}_{\mathrm{s}}]) - \mathrm{entropy}(\mathbb{E}[D^{(2)}_{\mathrm{s}}]) ... - \mathrm{entropy}(\mathbb{E}[D^{(g)}_{\mathrm{s}}])
\end{equation}
where $\mathbb{E}[\cdot]$ is computed along spatial dimensions while $\mathrm{entropy}(\cdot)$ is computed along the channel dimension.

\textit{Remark}. 
As illustrated in Fig.~\ref{fig:teaser}, by decomposing features into more reusable attributes, ideally any feature can be represented as a combination of these attributes, thus enhancing generalization. By indexing features with tuples rather than scalars, attribute-level similarities and differences can be captured for better object separability, thus benefiting convergence.
Notably, when $g$=1, the above formulation except Eq.~\ref{eq:utiliz_loss} reduces to the original non-grouped VAE.

However, directly grouping feature channels into different attributes may separate channels belonging to the same attribute apart or place channels belonging to different attributes together. This can degrade performance.

\begin{figure}[t]
\centering
\includegraphics[width=\linewidth]{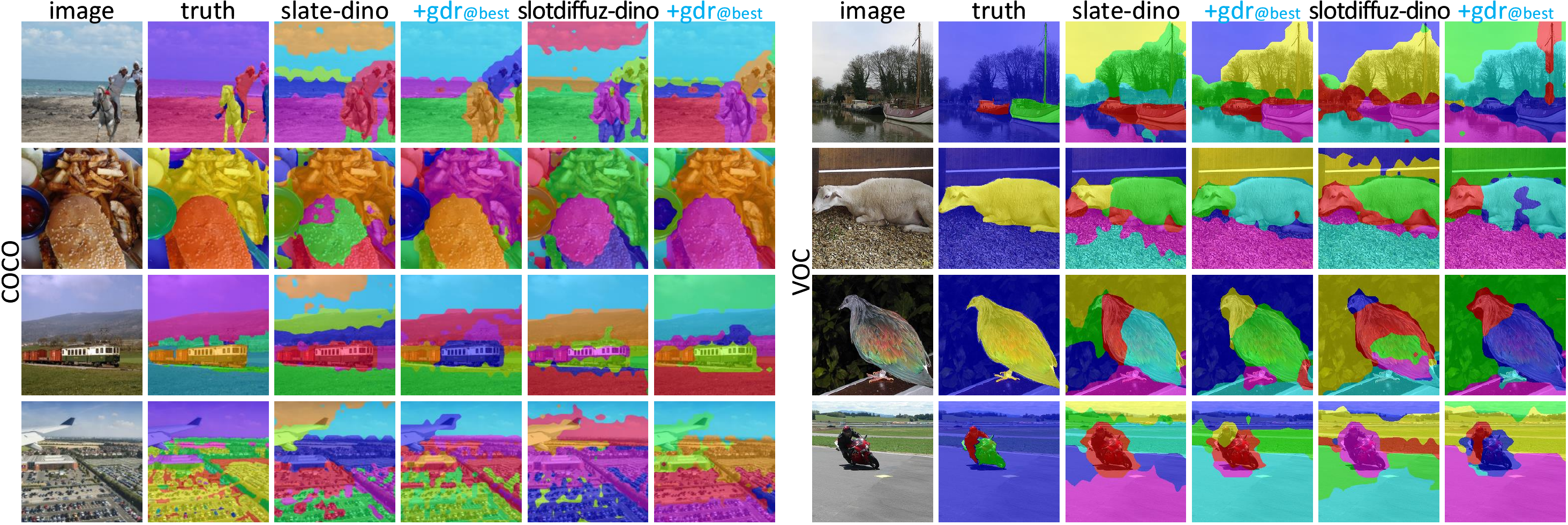}
\caption{
Object discovery visualization of SLATE and SlotDiffusion plus GDR.
}
\label{fig:ogdr_improves_tf_dm_dino_qualitative}
\end{figure}

\subsection{Organizing Channel Grouping}
\label{sect:ogdr}

In case incorrect channel grouping, we further design a channel organizing mechanism. The key idea is: We use an \textit{invertible projection} to organize the channel order of the continuous representation for grouped discretization, then apply this projection again to recover the (discretized) representation.

\textit{Firstly}, we project continuous representation $\bm{Z}$ to a higher channel dimension using the pseudo-inverse of a learnable matrix $\bm{W}$:
\begin{equation}
\label{eq:ogdr_project_up}
\small
\bm{Z}_+ = \bm{Z} \cdot \mathrm{pinv}(\bm{W})
\end{equation}
where $\bm{Z}$ is in shape (height, width, channel=$c$); $\mathrm{pinv}(\cdot)$ is pseudo-inverse; and matrix $\mathrm{pinv}(\bm{W})$ is in shape (channel=$c$, expanded channel=$8c$).
This facilitates channels belonging to the same attribute to be placed together by (\textit{i}) enabling channel reordering and (\textit{ii}) generating extra channels to mitigate mis-grouping.

\textit{Secondly}, we group $\bm{Z}_+$ along the channel dimension and discretize it using the attribute-level codebooks $\bm{C}$. This yields code indexes $\bm{X}_{\mathrm{i}}$ and the expanded discrete representation $\bm{X}_+$.
This follows the formulation in Eq.~\ref{eq:vqvae_dist}-\ref{eq:index_tuple_to_scalar} above.

\textit{Meanwhile}, we add $\bm{Z}_+$ to $\bm{X}_+$:
\begin{equation}
\label{eq:ogdr_residual}
\small
\bm{X}_+ := \bm{Z}_+ \times \alpha + \bm{X}_+ \times (1 - \alpha)
\end{equation}
where $\alpha$ decays via cosine annealing\footnote{\scriptsize https://pytorch.org/docs/stable/generated/torch.optim.lr\_scheduler.CosineAnnealingLR.html} from 0.5 to 0 in the first half of pretraining and is zeroed 0 afterward.
With such residual preserving information through the discretization, VAE can be well pretrained even under mis-grouping.

\textit{Thirdly}, we project $\bm{X}_+$ back to obtain the final organized grouped discrete representation:
\begin{equation}
\label{eq:ogdr_project_down}
\small
\bm{X} = \bm{X}_+ \cdot \bm{W}
\end{equation}
where $\bm{W}$ is the previously introduced learnable matrix in shape (expanded channel=8$c$, channel=$c$).

\textit{Fourthly}, to address potential numerical instability arising from matrix pseudo-inverse multiplcation, we normalize $\bm{X}$:
\begin{equation}
\label{eq:ogdr_normaliz}
\small
\bm{X} := \frac {\bm{X} - \mathbb{E}[\bm{X}]} {\sqrt{\mathbb{V}[\bm{X}] + \epsilon}}
\end{equation}
where $\mathbb{E}$ and $\mathbb{V}$ are the mean and variance over height, width and channel.

\begin{figure}[t]
\centering
\includegraphics[width=\linewidth]{figures/viz_pud.png}
\caption{
GDR's invertible projection learns to organize channels' orders for grouped discretization. Every sub-plot has three columns of channels (black bars) and matrix weights among them (grey ribbons). The first column corresponds to continuous representation channels. Ribbons between the first and second columns are the project-up weights. The second column is discretization attribute groups. Ribbons between the second and third columns are the project-down weights. The third column is discretized representation channels.
}
\label{fig:viz_pud}
\end{figure}

\subsection{{Grouped vs Non-Grouped}}
\label{sect:grouped_vs_nongrouped}

\textit{Codebook parameters}.
Compared to the non-grouped, the number of parameters in our grouped codebook is significantly reduced to $\frac{a g d}{a ^ g c}$=$\frac {a c} {a ^ g c}$=$\frac {1} {a ^ {g - 1}}$. E.g., only $\frac{1}{64}$ when $a$=64, $g$=2, $c$=256 and $a^g$=4096.
We increase $c$ to $8c$ and apply normalization plus linear to project it back to $c$, yielding $\frac{1}{1.6}$ the original number of codebook parameters -- still 30\% fewer.

\textit{Codebook computation}.
Non-grouped computation only involves code matching using inner product for each continuous feature: $c \times n \times 1$=$256 \times 4096$=$2 ^ {20}$.
GDR computation involves two projections and code matching: $8c \times c \times 2 + 8c \times \sqrt[g]{n}$, which results in $2^{20} + 2^{17}$ for $g$2 and $2^{20} + 2^{14}$ for $g$4 -- computation burden that is nearly identical to the original non-grouped case.

\begin{figure}[t]
\centering
\begin{subfigure}{0.495\textwidth}
\includegraphics[width=\textwidth]{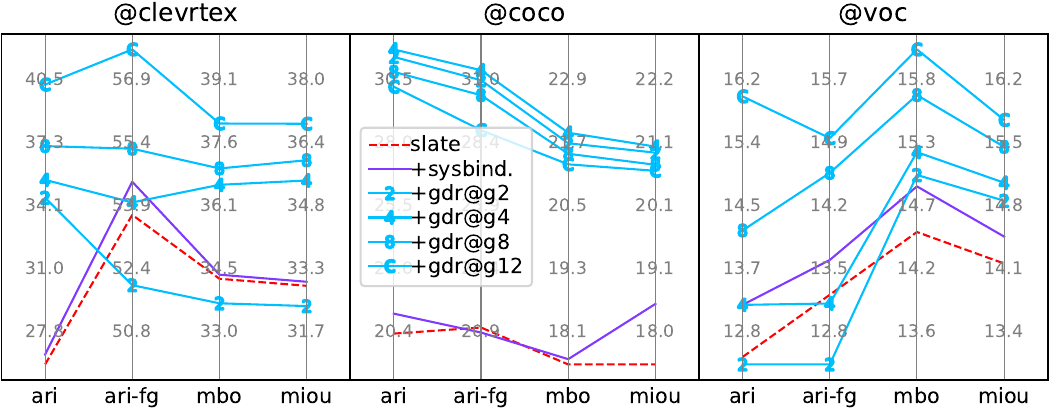}
\end{subfigure}
\begin{subfigure}{0.495\textwidth}
\includegraphics[width=\linewidth]{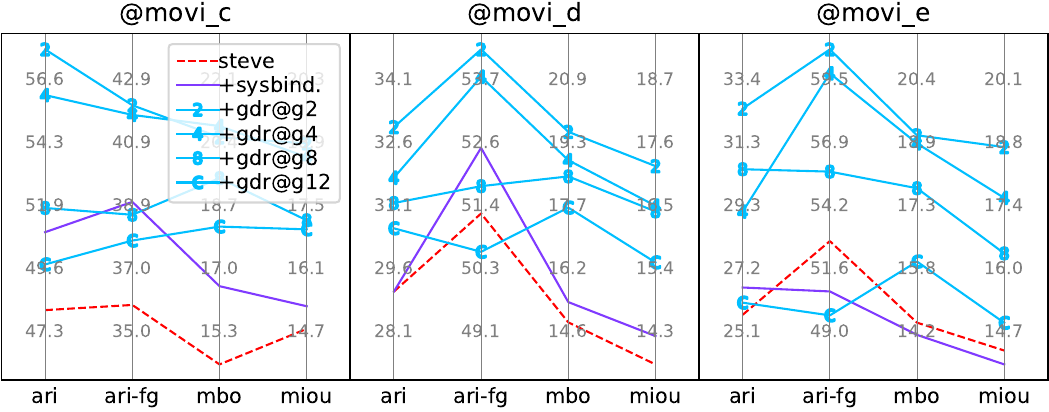}
\end{subfigure}
\begin{subfigure}{0.495\textwidth}
\includegraphics[width=\textwidth]{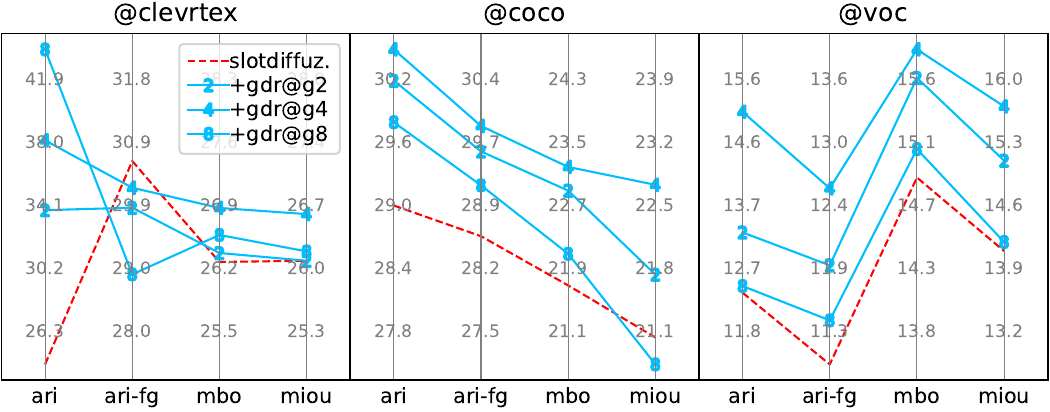}
\end{subfigure}
\begin{subfigure}{0.495\textwidth}
\includegraphics[width=\linewidth]{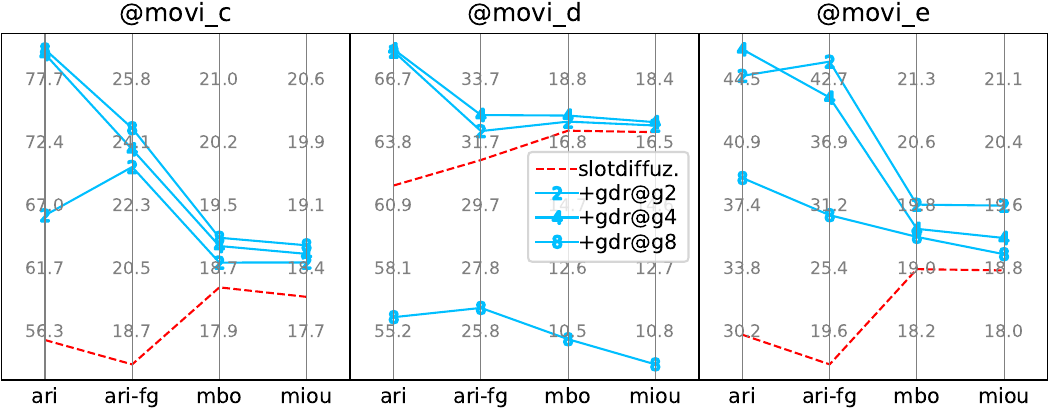}
\end{subfigure}
\caption{
GDR boosts object discovery performance of both Transformer- (\textit{top}) and Diffusion-based (\textit{bottom}) methods on images (\textit{left}) and videos (\textit{right}). A naive CNN is used as their primary encoder. Titles are datasets; x ticks are metrics while y ticks are metric values in adaptive scope. Higher values are better.
}
\label{fig:ogdr_improves_tf_dm}
\end{figure}

\begin{figure}[t]
\centering
\includegraphics[width=0.33\linewidth]{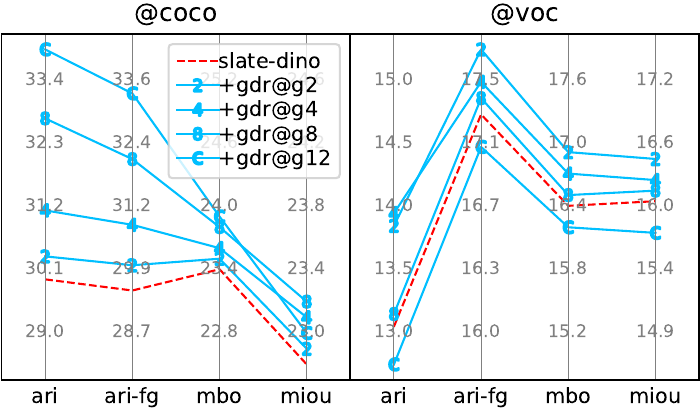}
\hspace{5em}
\includegraphics[width=0.33\linewidth]{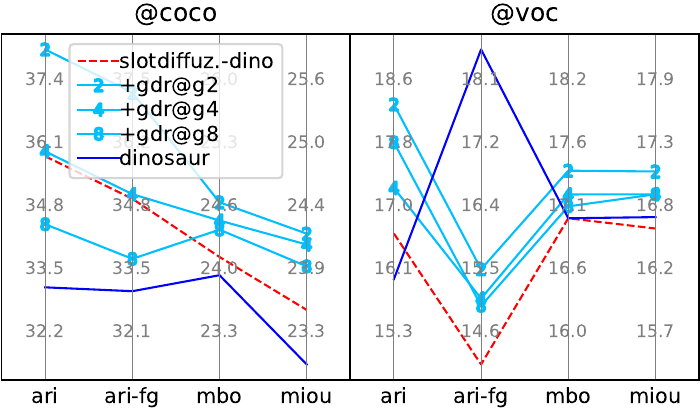}
\caption{
With DINO1-B/8 for primary encoding, GDR still improves Transformer- (\textit{left}) and Diffusion-based (\textit{right}) methods. Higher values are better.
}
\label{fig:ogdr_improves_tf_dm_dino}
\end{figure}

\section{Experiments}
\label{sect:experiments}

We conduct experiments using three random seeds to evaluate: (\textit{i}) How well GDR improves mainstream OCL, including Transformer- and Diffusion-based methods; (\textit{ii}) What visual intuitions GDR exhibits in VAE representation; (\textit{iii}) How designs of GDR contribute to its success in the OCL setting.

\subsection{{Experiment Overview}}
\label{sect:experiment_overview}

\begin{table}[t]
\centering\small
\setlength{\tabcolsep}{5pt}
\setlength{\aboverulesep}{0pt}  
\setlength{\belowrulesep}{0pt}  
\newcommand{\tss}[1]{\scalebox{0.8}{\texttt{#1}}}
\newcommand{\pv}[2]{#1\scalebox{0.4}{$\pm$#2}}
\newcommand{\cg}[1]{\textcolor{green}{#1}}
\newcommand{\ch}[1]{\textcolor{red}{#1}}

\begin{tabular}{cccccc}
\hline
COCO \tiny\#slots=7 & ARI\textsubscript{fg}$\uparrow$ & mBO$\uparrow$ 
& YTVIS \tiny\#slots=7 & ARI\textsubscript{fg}$\uparrow$ & mBO$\uparrow$ \\
\cmidrule(lr){1-3} \cmidrule(lr){4-6}
SPOT                & 37.5\psm{0.6} & 34.8\psm{0.1} 
& VideoSAUR           & 39.5\psm{0.6} & 29.0\psm{0.4} \\
SPOT+GDR@$g2$       & 39.7\psm{0.5} & 35.1\psm{0.1} 
& VideoSAUR+GDR@$g2$  & 43.6\psm{0.5} & 31.7\psm{0.4} \\
\hline
\end{tabular}

\vspace{0.5\baselineskip}
\setlength{\tabcolsep}{5pt}
\setlength{\aboverulesep}{0pt}  
\setlength{\belowrulesep}{0pt}  

\begin{tabular}{ccc}
\hline
COCO \tiny\#slots=7 & class@top1$\uparrow$  & bbox@R2$\uparrow$   \\
\cmidrule(lr){1-3}
SPOT + MLP & 0.59\psm{0.1} & 0.54\psm{0.1} \\
SPOT+GDR@$g$2 + MLP & 0.62\psm{0.1} & 0.56\psm{0.1} \\
\hline
\end{tabular}

\vspace{0.5\baselineskip}
\caption{Object discovery (\textit{upper}) and set prediction (\textit{lower}) of GDR upon state-of-the-arts, SPOT and VideoSAUR. DINO1-B/8 is used for primary encoding.}
\label{tab:spot_videosaur}
\end{table}

\textit{Models}.
We use both Transformer-based and Diffusion-based models as our GDR's basis. The former includes SLATE \cite{singh2021slate} for image and STEVE \cite{singh2022steve} for video. The latter includes SlotDiffusion \cite{wu2023slotdiffuz} and its temporal variant. Upon such basis, we compare GDR against SysBinder@$g$4 \cite{singh2022sysbind}.
We also apply GDR to state-of-the-art models, SPOT \cite{kakogeorgiou2024spot} and VideoSAUR \cite{zadaianchuk2024videosaur}, which are also Transformer-based.
Methods such as SA \cite{locatello2020slotattent} and SAVi \cite{kipf2021savi} are excluded due to their low performance or reliance on additional modalities.

\textit{Datasets}.
We evaluate those models on ClevrTex\footnote{\scriptsize https://www.robots.ox.ac.uk/\raisebox{-0.25em}{\textasciitilde}vgg/data/clevrtex}, COCO\footnote{\scriptsize https://cocodataset.org/\#panoptic-2020} and VOC\footnote{\scriptsize http://host.robots.ox.ac.uk/pascal/VOC} for image OCL tasks, while MOVi-C/D/E\footnote{\scriptsize https://github.com/google-research/kubric/tree/main/challenges/movi} for video. We also use YTVIS\footnote{\scriptsize https://youtube-vos.org/dataset/vis}, YouTube video instance segmentation. These encompass both synthetic and real-world cases, featuring multiple objects and complex textures. Except for those two state-of-the-arts, the input size is unified to 128$\times$128 and other data processing follows the convention. Note that we use COCO panoptic instead of instance segmentation and the high-quality YTVIS\footnote{\scriptsize https://github.com/SysCV/vmt?tab=readme-ov-file\#hq-ytvis-high-quality-video-instance-segmentation-dataset} for strict evaluation.

\textit{Hyperparameters}.
The codebook size is $n$=$a^g$=4096 for all. GDR's group number is set to GDR@$g$2, $g$4, $g$8 and $g$12. Correspondingly, the attribute group sizes are (64, 64), (8, 8, 8, 8), (2, 2, 2, 2, 4, 4, 4, 4) and (2, 2, 2, 2, 2, 2, 2, 2, 2, 2, 2, 2), ensuring 4096 combinations.
The number of slots is roughly set to the maximum/average object count plus one: 10+1 for ClevrTex, COCO, VOC and MOVi-C; 20+1 for MOVi-D; and 23+1 for MOVi-E. However, for those two state-of-the-arts, we strictly follow their official experiment settings.

\subsection{Performance}
\label{sect:performance}

\begin{figure}[t]
\centering
\includegraphics[width=0.49\textwidth]{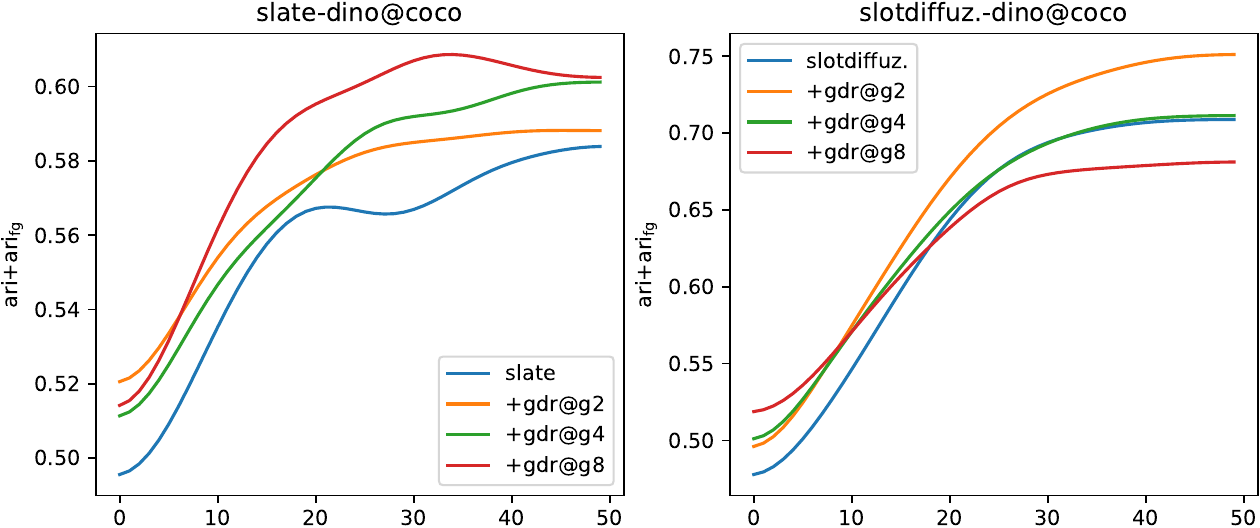}
\hspace{3em}
\includegraphics[width=0.33\textwidth]{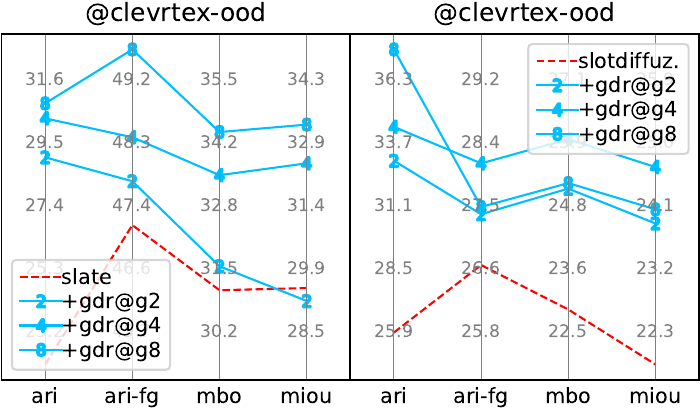}
\caption{
(\textit{left}) GDR accelerates model convergence. The x axis is val epochs while y is accuracy in ARI+ARI\textsubscript{fg}. Smoothed with a Gaussian kernel size 5. 
(\textit{right}) GDR improves model generalization. Models are trained on Clevrtex and tested on its out-of-distribution version. Higher values are better.
}
\label{fig:converge}
\end{figure}

\textit{Object discovery}. 
We use common object discovery metrics: Adjusted Rand Index (ARI)\footnote{\scriptsize https://scikit-learn.org/stable/modules/generated/sklearn.metrics.adjusted\_rand\_score.html}, ARI foreground (ARI\textsubscript{fg}), mean Best Overlap (mBO)\footnote{\scriptsize https://ieeexplore.ieee.org/document/7423791} and mean Intersection-over-Union (mIoU)\footnote{\scriptsize https://scikit-learn.org/stable/modules/generated/sklearn.metrics.jaccard\_score.html}.
As shown in Fig.~\ref{fig:ogdr_improves_tf_dm}, GDR significantly enhances accuracy across both synthetic and real-world images and videos. With the naive CNN \cite{kipf2021savi} for primary encoding, it boosts both Transformer-based methods and Diffusion-based methods. GDR always outperforms the competitor SysBinder by a large margin.
We further evaluate GDR's effectiveness with vision foundation model DINO1-B/8 \cite{caron2021dino1} for strong primary encoding. As shown in Fig.~\ref{fig:ogdr_improves_tf_dm_dino} and \ref{fig:ogdr_improves_tf_dm_dino_qualitative}, on both SLATE and SlotDiffusion, GDR improves accuracy across all metrics in most cases. 

\textit{Applying to state-of-the-art}. Following the original settings, we apply GDR to SPOT \cite{kakogeorgiou2024spot} and VideoSAUR \cite{zadaianchuk2024videosaur}, by replacing SPOT's VAE with GDR and by replacing VideoSAUR's reconstruction target (continuous DINO features) with GDR discretized DINO features, respectively. As shown in Table \ref{tab:spot_videosaur} upper, GDR is still able to boost state-of-the-art methods' performance further.

\textit{Set prediction}.
Following \cite{seitzer2023dinosaur}, we employ OCL to represent dataset COCO as slots, and use a small MLP to predict the object class and bounding box corresponding to each slot, with metrics of top-1 accuracy and the R2 score respectively. As shown in Table \ref{tab:spot_videosaur} lower, our GDR improves SLATE in set prediction, demonstrating is superior quality in object representation.

\begin{figure}[t]
\centering
\includegraphics[width=\linewidth]{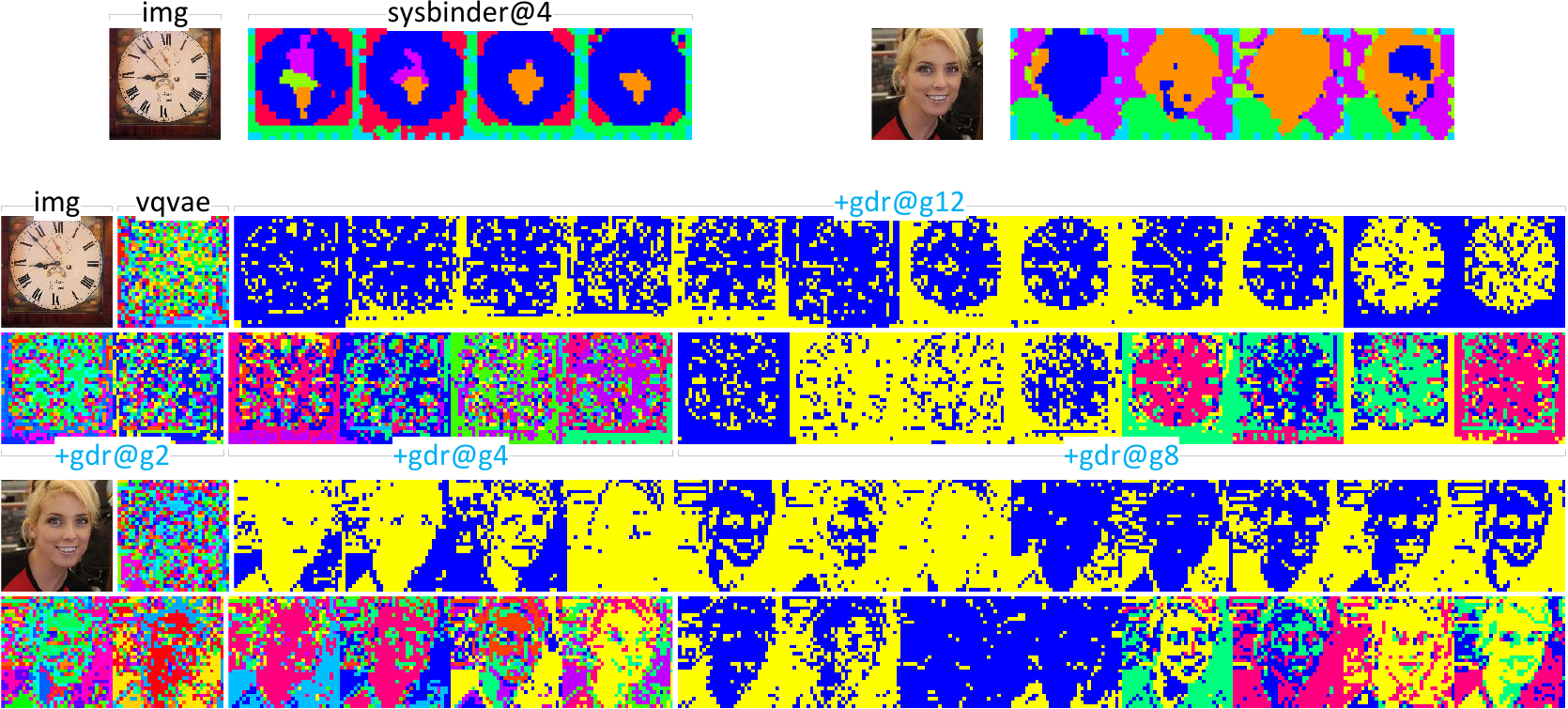}
\caption{
SysBinder's attribute groups (\textit{upper}), i.e., attention maps, and GDR's attribute groups (\textit{lower}), i.e., tuple code indexes.
GDR captures attribute-level similarities and differences among superpixels, whereas the non-grouped ``vqvae'' mixes all together. SysBinder lacks too much diversity and detail.
}
\label{fig:viz_zidx}
\end{figure}

\textit{Convergence}. 
The validation curves of ARI+ARI\textsubscript{fg} in Fig.~\ref{fig:converge} left demonstrate that GDR consistently accelerates the basis' convergence in OCL training. Along with Fig.~\ref{fig:viz_zidx}, forming VAE discrete representation with tuple indexes captures attribute-level similarities and differences among super-pixels, thereby guiding OCL models to converge better.

\textit{Generalization}. 
We transfer models from ClevrTex to its out-of-distribution version ClevrTex-OOD without finetuning. As shown in Fig.~\ref{fig:converge} right, GDR consistently improves basis methods' generalization. This confirms that GDR's decomposition from features into attributes helps the model learn more fundamental representations that are robust to distribution shifts.

\begin{figure}[t]
\centering
\includegraphics[width=1\textwidth]{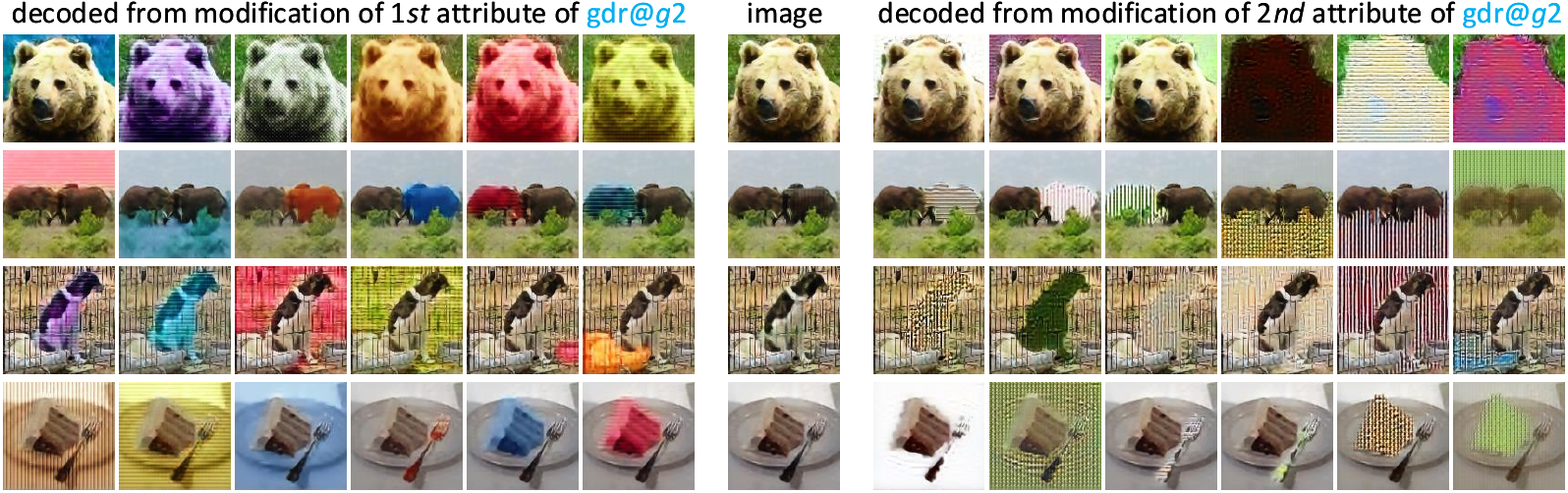}
\caption{
For GDR@$g$2, one attribute group roughly learns colors, while the other roughly learns textures.
The original image is at the center. The left and right are images decoded from the modified VAE discrete representation.
}
\label{fig:visualiz_idx_dec}
\end{figure}

\begin{figure}[t]
\centering
\includegraphics[width=0.95\linewidth]{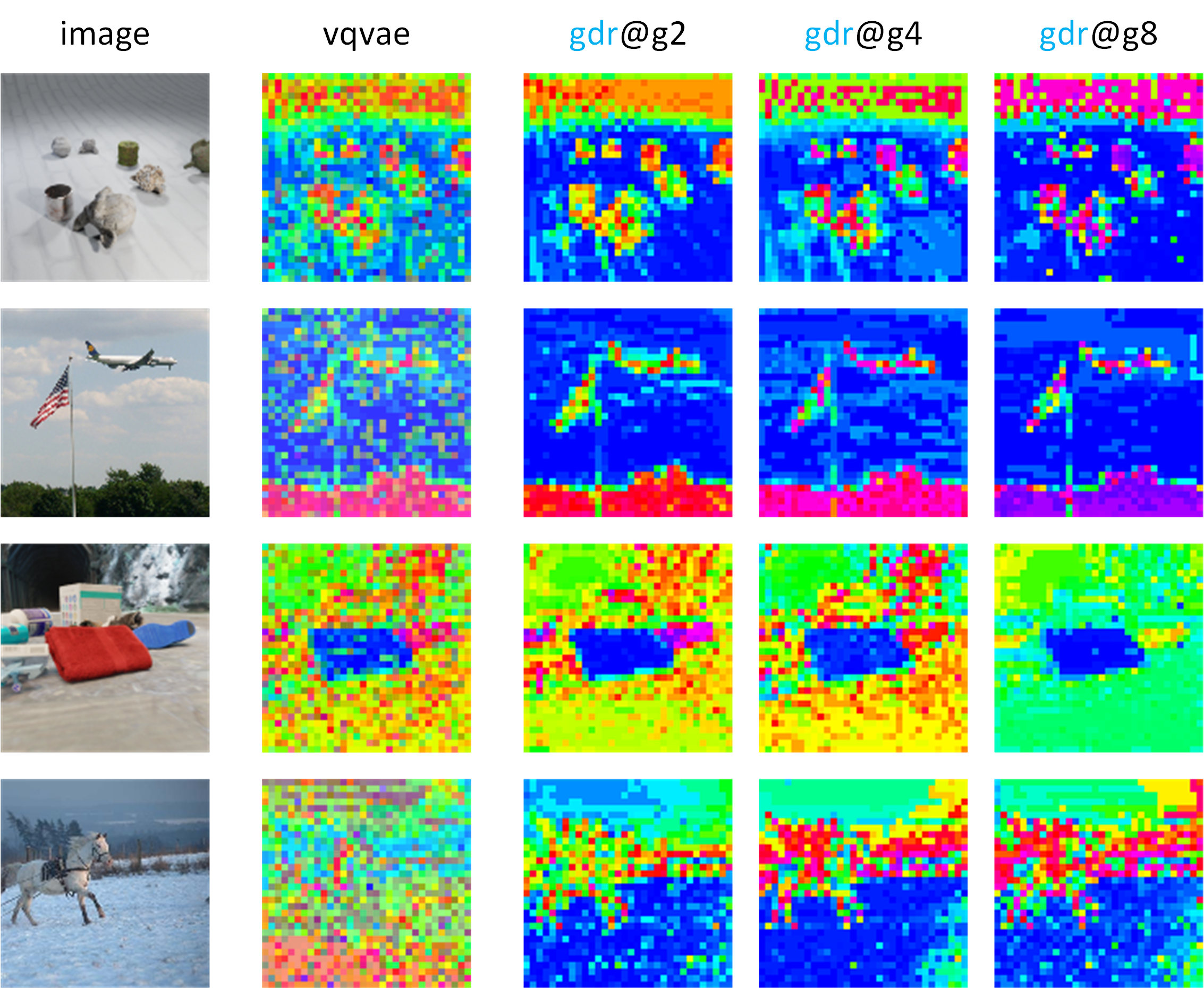}
\caption{
GDR improves object separability in VAE discrete representation.
More groups improve object separability but increase the risk of losing some objects.
}
\label{fig:object_separability}
\end{figure}

\subsection{Interpretability}
\label{sect:interpretability}

\textit{Decomposition from features to attributes}. 
Although without explicit supervision models can hardly learn concepts \cite{singh2022sysbind} as human-readable as in Fig.~\ref{fig:teaser}, we can still analyze GDR's decomposition from features to attributes as follows.
Given GDR discrete representation $\bm{X}$, we replace the attributes of objects' superpixels with arbitrary attribute codes then decode them into images to observe the changes.
As shown in Fig.~\ref{fig:visualiz_idx_dec}, under $g$2 setting, modifying one attribute group roughly alters the colors, whereas modifying the other destroys the textures. This suggests that the first group learns colors while the second learns textures.

\textit{Attribute-level similarities and differences}. 
Basis methods' scalar index tensor and GDR's tuple index tensor $\bm{X}_{\mathrm{i}}$ can be visualized by mapping different indexes to distinct colors. 
For our only competitor, SysBinder, we assign different colors to its different attention groups.
As shown in Fig.~\ref{fig:viz_zidx}, scalar indexes mix all attributes together, whereas our tuple indexes highlight similarities (identical colors) and differences (distinct colors) among superpixels in each attribute group.
In contrast, SysBinder also captures such attribute-level information but with very limited diversity and details.

\textit{Object separability}. 
The visualization of $\bm{X}$ for both the basis VQ-VAE and GDR can be achieved by coloring the different distances between each superpixel and the reference point (the average of all superpixels).
As shown in Fig.~\ref{fig:object_separability}, GDR consistently exhibits better object separability across all $g$ settings, suggesting GDR's superior guidance to OCL. However, using an excessive number of groups in GDR may result in the omission of certain objects.

\subsection{Ablation}
\label{sect:ablation}

The effects of different designs in GDR are listed in Tab.~\ref{tab:ablation}. We use ARI+ARI\textsubscript{fg} since ARI largely indicates how well the background is segmented while ARI\textsubscript{fg} reflects the discovery quality of foreground objects.

\textit{Number of groups}, formulated in Eq.~\ref{eq:vqvae_dist}-\ref{eq:utiliz_loss}: $g$=2, 4, 8 or 12. As shown in Fig.~\ref{fig:ogdr_improves_tf_dm} and \ref{fig:ogdr_improves_tf_dm_dino}, the optimal $g$ depends on the specific dataset. However, $g$12 and $g$8 tend to result in suboptimal performance while $g$4 consistently leads to guaranteed performance gains over the basis methods.

\textit{Channel expansion rate}, formulated in Eq.~\ref{eq:ogdr_project_up}: $c$, 2$c$, 4$c$ or 8$c$. Although $8c$ generally performs best, the expansion rate has a nearly saturated impact on GDR's performance. This suggests that our channel organizing mechanism is effective, reducing the necessity for a higher channel expansion rate.

The \textit{channel organizing} based on our \textit{invertible projection} designed in Sect.~\ref{sect:ogdr} is crucial for OCL model performance. If we disrupt it by replacing $W$ pseudo-inverse in project-up with specified weights, as formulated in Eq. \ref{eq:ogdr_project_up}, the object discovery accuracy drops significantly.

We also visualize how the invertible project-up and project-down organize channels for grouping. As shown in Fig.~\ref{fig:viz_pud}, some input channels are mixed, switched or split into different attributes for discretization, then the pseudo-inverse recovers them in the form of discrete representations. Such patterns are clearly observed across most datasets and grouping configurations.

Using \textit{annealing residual connection} during training, formulated in Eq.~\ref{eq:ogdr_residual}, consistently yields better performance than without.

\textit{Normalization at last}, formulated in Eq.~\ref{eq:ogdr_normaliz}, is generally beneficial, though its effect is not highly significant.

\begin{table}[t]
\centering\small
\setlength{\tabcolsep}{5pt}
\setlength{\aboverulesep}{0pt}  
\setlength{\belowrulesep}{0pt}  

\begin{tabular}{ccccc}
\hline
expansion rate & 8$c$  & 4$c$  & 2$c$  & 1$c$  \\
\cmidrule(lr){1-1} \cmidrule(lr){2-5}
ARI+ARI\textsubscript{fg}   & 89.47 & 89.29 & 88.93 & 88.16 \\
\hline
\end{tabular}
\\
\makecell{}
\\
\begin{tabular}
{ccccc}
\hline
\textls[-50]{utiliz. loss}
& \textls[-50]{invertible project}
& \textls[-50]{residual connection}
& \textls[-50]{final normaliz.}
& ARI+ARI\textsubscript{fg} \\
\cmidrule(lr){1-4} \cmidrule(lr){5-5}
\ding{51}   & \ding{51}   & \ding{51}   & \ding{51} & 89.47 \\
\arrayrulecolor[rgb]{0.9,0.9,0.9}
\cmidrule(lr){1-4} \cmidrule(lr){5-5}
\ding{55}   &             &             &           & 84.78 \\
\cmidrule(lr){1-4} \cmidrule(lr){5-5}
            & \ding{55}   &             &           & 81.52 \\
\cmidrule(lr){1-4} \cmidrule(lr){5-5}
            & \ding{55} $W$pinv &       &           & 32.25 \\
\cmidrule(lr){1-4} \cmidrule(lr){5-5}
            &             & \ding{55}   &           & 88.84 \\
\cmidrule(lr){1-4} \cmidrule(lr){5-5}
            &             &             & \ding{55} & 89.16 \\
\arrayrulecolor{black}
\hline
\end{tabular}

\vspace{0.5\baselineskip}
\caption{Effects of expansion rate, utilization loss, invertible projection (and replacing $W$ pinv with specified weights), residual connection in training and final normalization. Experimented SLATE+GDR@$g$4 on ClevrTex.
}
\label{tab:ablation}
\end{table}

\section{Conclusion}
\label{sect:conclusion}

We propose grouped discrete representation in VAE to guide OCL training better. This technique improves the mainstream Transformer- and Diffusion-based OCL methods in both convergence and generalization. Although self-supervision cannot guarantee different groups learn different human-readable attributes, our method still exhibits interesting and interpretable patterns in attribute-level discrete representations. Fundamentally, we only modify the VAE part of OCL models, indicating broader applicability to other VAE-based models.

\section*{Acknowledgment}

We acknowledge the support of the Finnish Center for Artificial Intelligence (FCAI) and the Research Council of Finland through its Flagship program. 
Additionally, we thank the Research Council of Finland for funding the projects ADEREHA (grant no. 353198), BERMUDA (362407) and PROFI7 (352788).
We also appreciate CSC-IT Center for Science, Finland, for granting access to the LUMI supercomputer, owned by the EuroHPC Joint Undertaking and hosted by CSC (Finland) in collaboration with the LUMI consortium. 
Furthermore, we acknowledge the computational resources provided by the Aalto Science-IT project through the Triton cluster.
Finally, the first author expresses his heartfelt gratitude to his wife for her unwavering support and companionship.

\bibliographystyle{splncs04}
\bibliography{paper}

\end{document}